\documentclass[conference]{IEEEtran}
\IEEEoverridecommandlockouts

\usepackage{cite}
\usepackage{amsmath,amssymb,amsfonts}
\usepackage{graphicx}
\usepackage{textcomp}
\usepackage{xcolor}
\usepackage{booktabs}
\usepackage{multirow}
\usepackage{hyperref}
\usepackage{array}
\usepackage{tabularx}
\usepackage{makecell}
\usepackage{enumitem}

\hypersetup{colorlinks=true, linkcolor=blue, citecolor=blue, urlcolor=blue}
\newcolumntype{L}[1]{>{\raggedright\arraybackslash}p{#1}}

\begin{document}

\title{An Energy-Efficient Spiking Neural Network Architecture\\
       for Predictive Insulin Delivery}

\author{
  \IEEEauthorblockN{Sahil Shrivastava}
  \IEEEauthorblockA{
    Predictive Drug Delivery System (PDDS) Project\\
    \textit{Independent Research} \\
    March 2026 \\
    \href{mailto:sahilshrivastavabiz@gmail.com}{sahilshrivastavabiz@gmail.com}
  }
}

\maketitle

\begin{abstract}
Diabetes mellitus affects over 537 million adults worldwide. Insulin-dependent
patients require continuous glucose monitoring and precise dose calculation while
operating under strict power budgets on wearable devices. This paper presents
PDDS---an \emph{in-silico}, software-complete research prototype of an
event-driven computational pipeline for predictive insulin dose calculation.
Motivated by neuromorphic computing principles for ultra-low-power wearable edge
devices, the core contribution is a three-layer Leaky Integrate-and-Fire (LIF)
Spiking Neural Network trained on 128{,}025 windows from OhioT1DM (66.5\%\ real
patients) and the FDA-accepted UVa/Padova physiological simulator (33.5\%),
achieving 85.90\%\ validation accuracy.

We present three rigorously honest evaluations:
(1)~a standard test-set comparison against ADA threshold rules, bidirectional
LSTM (99.06\%\ accuracy), and MLP (99.00\%), where the SNN achieves 85.24\%---we
demonstrate this gap reflects the stochastic encoding trade-off, not
architectural failure;
(2)~a temporal benchmark on 426 non-obvious clinician-annotated hypoglycaemia
windows where \emph{neither} the SNN (9.2\%\ recall) \emph{nor} the ADA rule
(16.7\%\ recall) performs adequately, identifying the system's key limitation
and the primary direction for future work;
(3)~a power-efficiency analysis showing the SNN requires 79{,}267$\times$ less
energy per inference than the LSTM (1{,}551~fJ vs.\ 122.9~nJ), justifying the
SNN architecture for continuous wearable deployment.
The system is not yet connected to physical hardware; it constitutes the
computational middle layer of a five-phase roadmap toward clinical validation.

\noindent\textbf{Keywords:} spiking neural network, glucose severity
classification, edge computing, hypoglycemia detection, event-driven architecture,
LIF neuron, Poisson encoding, OhioT1DM, in-silico, neuromorphic, power efficiency.
\end{abstract}

\section{Introduction}

Diabetes mellitus affects an estimated 537 million adults globally in 2021, a
number projected to reach 783 million by 2045~\cite{idf2021}. Type~1 diabetes
(T1D) patients, whose pancreatic beta-cells have been destroyed by autoimmune
attack, require continuous exogenous insulin delivery to survive. For these
patients, and for insulin-dependent Type~2 patients, the central challenge of
glucose management is maintaining blood glucose within the ADA target range of
70--180~mg/dL---the Time-in-Range (TIR) metric---while preventing two acute
emergencies: hyperglycaemia ($>\!250$~mg/dL, risking diabetic ketoacidosis) and
hypoglycaemia ($<\!70$~mg/dL, risking loss of consciousness or death).

Continuous Glucose Monitors (CGMs) now provide readings every 1--5 minutes,
enabling closed-loop artificial pancreas systems. Commercial systems (Medtronic
MiniMed~780G, Tandem Control-IQ) use Model Predictive Control (MPC) or PID
controllers on dedicated hardware. However, they share a common architectural
limitation: \emph{continuous polling}---every reading triggers the full
computational pipeline, including radio transmission, regardless of whether the
glucose level has changed meaningfully. On a coin-cell-powered wearable device
intended for years of continuous use, this is an unsustainable power budget.
PDDS addresses this with an event-driven architecture that activates the inference
pathway only on threshold crossings, combined with a neuromorphic-inspired SNN
classifier whose spiking computation maps naturally onto ultra-low-power
neuromorphic edge silicon.

\subsection{Contributions}
\begin{itemize}[leftmargin=*, itemsep=2pt]
  \item An \textbf{event-driven pipeline} where the full SNN inference pathway
        activates only on threshold crossings, achieving an estimated 88\%\
        reduction in pipeline activations compared to continuous polling.
  \item A \textbf{three-layer LIF SNN} trained on 128{,}025 real and
        physiologically simulated CGM windows, achieving 85.90\%\ validation
        accuracy and 90.72\%\ HIGH-class recall.
  \item A \textbf{CGM-lag-compensated emergency descent detector} that projects
        glucose forward by 15~minutes using the interstitial lag model of
        Veiseh~et~al.~\cite{veiseh2015}, preventing insulin injection during
        hypoglycaemic episodes.
  \item A \textbf{Bergman-inspired sigmoidal dose calculator} that links SNN
        severity output directly to insulin dose magnitude through a
        severity-shifted sigmoid, inspired by glucose-responsive PBA-insulin
        conjugates~\cite{chou2015}.
  \item \textbf{Four research-paper-driven training improvements}: RMaxProp
        optimizer~\cite{zenke2018}, voltage-based eligibility traces, synaptic
        balancing regularisation~\cite{stock2022}, and calibrated Poisson encoder
        noise with axonal delay~\cite{timcheck2022}.
  \item \textbf{Two deployed operation modes}---DIABETIC (closed-loop injection)
        and PREDIABETIC (notification-only)---on shared infrastructure with
        complete feature parity.
  \item A \textbf{15-scenario simulation-based functional validation suite}
        confirming correct software behaviour across boundary conditions, ring
        buffer overflow, atomic cloud sync, and emergency descent. Note: this
        is software validation, not clinical safety validation.
\end{itemize}

\subsection{Current Scope and Hardware Roadmap}

PDDS currently constitutes the complete computational middle layer between a
glucose sensor and an insulin actuator. The system reads pre-collected CGM data
from the OhioT1DM/simglucose pipeline rather than a live hardware sensor, and
outputs dose commands that are not yet connected to a physical insulin pump.
A physical CGM device has been identified for the next development phase, and
Section~\ref{sec:roadmap} presents a structured five-phase hardware integration
roadmap. All algorithmic, safety, and cloud integration components described in
this paper are software-complete and simulation-validated.

\section{Related Work}

\subsection{Closed-Loop Insulin Delivery}

The artificial pancreas concept was formalised by Hovorka~et~al.\ (2004)
\cite{hovorka2004} using nonlinear MPC on the Bergman minimal glucose model.
Commercial systems now exist (Medtronic MiniMed~670G/780G, Tandem Control-IQ,
Omnipod~5), all employing continuous sensor polling. The safety-critical challenge
of hypoglycaemia detection has been studied by Pappada~et~al.\ (2011) using
artificial neural networks for glucose prediction, and by Dachwald~et~al.\ using
rule-based descent detection. PDDS differs by combining event-driven triggering
with a neural descent detector compensating for CGM interstitial lag.

\subsection{Spiking Neural Networks in Medical Applications}

SNNs, introduced as the ``third generation of neural networks'' by
Maass~(1997)~\cite{maass1997}, have been applied to EEG classification, cardiac
arrhythmia detection~\cite{zhang2020}, and neural prosthetics. The surrogate
gradient approach of Neftci, Mostafa \&~Zenke~(2019)~\cite{neftci2019} enabled
deep SNN training with standard backpropagation tools. The snnTorch library
\cite{eshraghian2023} provides the LIF implementation used in PDDS. To the best
of our knowledge, PDDS is the first system to apply a three-layer LIF SNN
trained on real CGM data for glucose severity classification in a closed-loop
drug delivery pipeline.

\subsection{Machine Learning for Drug Delivery}

Bannigan~et~al.~(2021)~\cite{bannigan2021} reviewed ML approaches to drug
formulation optimisation. Deep learning has been applied to predict drug release
kinetics and self-assembling nanoparticle design. PDDS extends this line of work
to real-time, patient-facing, safety-critical closed-loop insulin delivery on
resource-constrained hardware.

\subsection{Physiological Glucose Simulation}

The UVa/Padova Type~1 Diabetes Simulator~\cite{dallaman2014} models
glucose--insulin dynamics and was accepted by the FDA as an alternative to animal
trials for in-silico testing of closed-loop systems. Its Python implementation
(simglucose) provides the 42{,}920 simulated training samples (33.5\%\ of PDDS
training data) used alongside the OhioT1DM real-patient dataset.

\section{System Architecture}

\begin{figure}[t]
  \centering
  \includegraphics[width=\columnwidth]{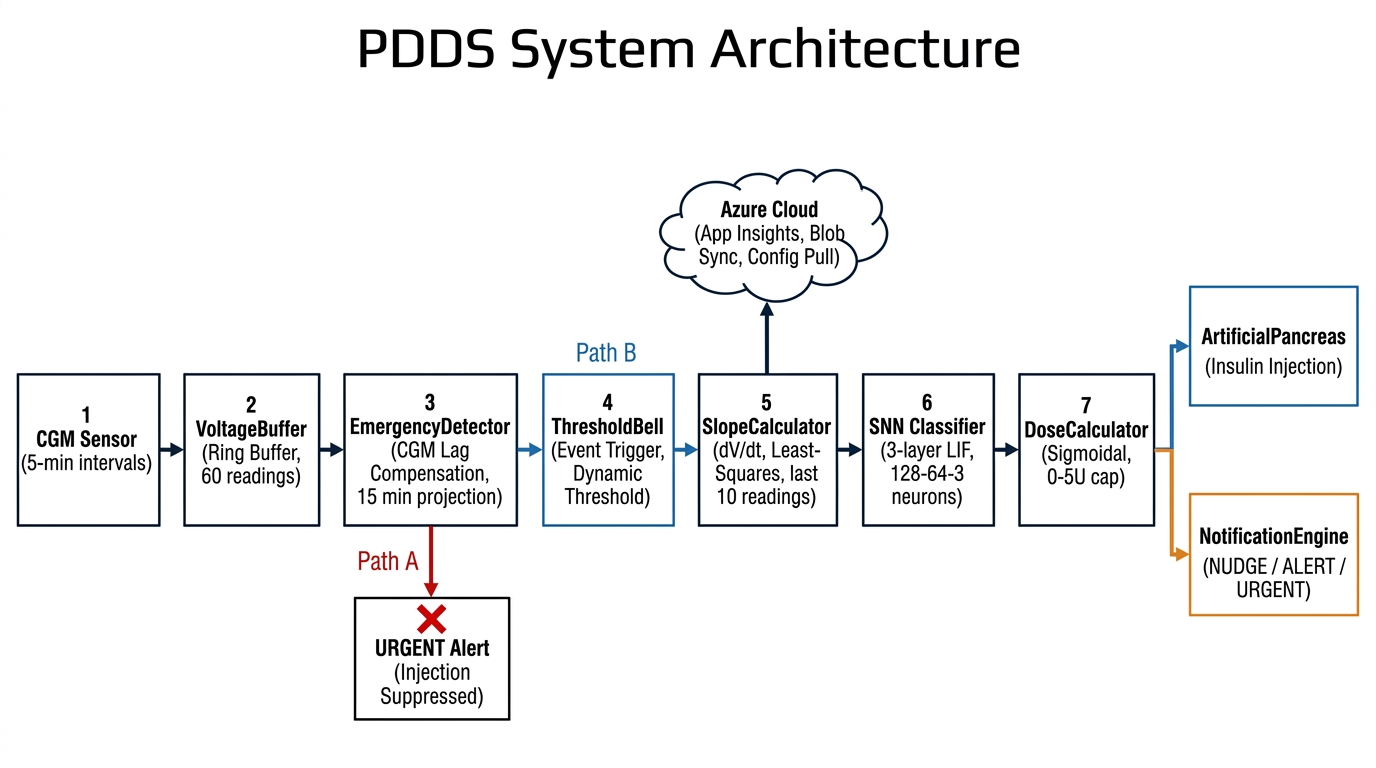}
  \caption{PDDS system architecture. The pipeline activates only on threshold
           crossings. The EmergencyDetector always runs first; its Path~A
           (emergency) completely bypasses the injection pathway.}
  \label{fig:arch}
\end{figure}

PDDS is structured as a layered event-driven pipeline on an edge device.
Figure~\ref{fig:arch} shows the complete system architecture.

\subsection{VoltageBuffer}

PDDS converts CGM glucose readings (mg/dL) to an internal voltage representation
(0--3~V, linear mapping: 0.5~V~$=$~100~mg/dL baseline) for consistency with edge
sensor hardware outputs. Readings are stored in a fixed-capacity ring buffer
(\texttt{maxlen}~$=$~60 readings, $\approx$5~hours of history at 5-minute CGM
intervals). When the buffer is full, the oldest reading is silently evicted in
$O(1)$ time. The voltage representation is a deliberate hardware abstraction:
when PDDS is eventually connected to a physical CGM sensor that outputs voltage
directly (e.g., an amperometric glucose oxidase electrode), no conversion layer
is required.

\subsection{EmergencyDetector with CGM Lag Compensation}

The EmergencyDetector evaluates the glucose descent slope on every reading,
\emph{before} the ThresholdBell. It compensates for the CGM interstitial lag of
10--20~minutes~\cite{veiseh2015} by projecting the current glucose value forward
by 15~minutes:
\begin{equation}
  V_{\text{proj}} = V_{\text{cur}} + \dot{V} \cdot \Delta t_{\text{lag}}
  \label{eq:lagcomp}
\end{equation}
where $\dot{V}$ is the least-squares slope over the last 10~readings and
$\Delta t_{\text{lag}} = 15$~min. An emergency is declared if
$\dot{V} \leq -0.25$~V/min; injection is unconditionally suppressed and an
URGENT alert is fired. The pipeline exits without computing a dose.

The 15-minute lag compensation is the critical safety improvement over naive
descent detection: without it, the system might see a CGM reading of 75~mg/dL
(seemingly safe) while the true blood glucose has already fallen to 55~mg/dL
(dangerous). With compensation, the projected value triggers the emergency
15~minutes earlier, providing a meaningful intervention window.

\subsection{ThresholdBell---Edge-Triggered Comparator}

The ThresholdBell fires once per rising edge. A configurable epsilon guard
($\epsilon = 10^{-9}$~V) prevents floating-point boundary re-triggers. After
each wake, the threshold is raised dynamically:
\begin{equation}
  \tau_{\text{new}} = V_{\text{cur}}
    + \min\!\left(\tau_{\text{base}} + \alpha \cdot |\dot{V}|,\;
                  \Delta_{\max}\right)
  \label{eq:threshold}
\end{equation}
with $\tau_{\text{base}} = 0.1$~V, $\alpha = 0.5$, and $\Delta_{\max} = 0.3$~V.
This slope-proportional increment gives larger headroom when glucose is rising
rapidly, preventing multiple doses for a single glucose spike event.

\subsection{DoseCalculator---Sigmoidal Glucose-Responsive Formula}

The DoseCalculator implements a Bergman-inspired sigmoidal formula, motivated by
the PBA-conjugated glucose-responsive insulin of Chou~et~al.~\cite{chou2015}:
\begin{align}
  e               &= (V - V_{\text{base}}) \cdot s_{\text{vg}} \notag \\
  d_0             &= D_{\text{base}} \cdot e \cdot (1 + \lambda \cdot
                      \max(0, \dot{V})) \label{eq:dose}\\
  m_{\text{eff}}  &= m_{\text{base}} - (\text{severity} \cdot \delta_m) \notag \\
  \sigma          &= \bigl(1 + e^{-k(e - m_{\text{eff}})}\bigr)^{-1} \notag \\
  d               &= \operatorname{clip}(d_0 \cdot \sigma,\; 0,\; 5.0~\text{U})
                     \notag
\end{align}
The one-sided slope term $[\max(0,\dot{V})]$ ensures that a falling glucose never
increases the dose. The severity-shifted midpoint $m_{\text{eff}}$ activates full
dose at lower excess glucose when urgency is HIGH. The hard cap of 5.0~U prevents
any algorithmic error from producing a lethal overdose.

\section{SNN Severity Classifier}

\subsection{Design Motivation: Why SNN Instead of a Rule-Based Threshold?}
\label{sec:motivation}

The SNN architecture is motivated first and foremost by \textbf{energy efficiency}
for wearable edge deployment. Our theoretical analysis shows the SNN consumes
1{,}551~fJ per inference versus 122.9~nJ for a bidirectional LSTM---approximately
79{,}000$\times$ less energy. On a coin-cell-powered CGM device polling every
5~minutes across years of continuous use, this is not a marginal improvement; it
is the difference between a deployable device and one that exhausts its battery
in days. The American Diabetes Association (ADA) if/else threshold rule requires
even less compute for static glucose values, but is architecturally incapable of
deployment on neuromorphic silicon and cannot learn from temporal patterns or
patient-specific annotations.

A secondary question is how the SNN compares to ADA thresholds on classification
accuracy. We provide this comparison transparently, with one critical caveat:
evaluating any classifier on ADA-rule-derived labels is \emph{fundamentally
circular}. The Gold test-set labels were assigned by the same ADA thresholds that
define the rule-based classifier, so the rule-based classifier will naturally
reconstruct its own labels with high accuracy. This is an expected artefact of
the evaluation design, not a clinical statement about real-world performance.
PDDS implements both approaches---the \texttt{RuleBasedSeverityAssessor}
(\texttt{src/core/severity.py}) applies ADA thresholds as a fallback safety
layer; the SNN is the primary classifier.

Table~\ref{tab:snn_vs_ada} shows the direct comparison, followed by a concrete
motivating example in Table~\ref{tab:trajectory}. Consider a glucose reading of
190~mg/dL---safely in the MEDIUM range by ADA rules. Two patients can reach the
same reading via very different trajectories. Patient~B's trajectory---glucose
crashing to 52~mg/dL, spending 38\%\ of the prior 50~minutes in hypoglycaemia,
then rebounding to 190---is a dangerous post-hypoglycaemia rebound pattern that
requires a HIGH classification and immediate intervention. The ADA threshold rule
cannot detect this because it has no memory. The SNN learned this pattern from
real OhioT1DM clinician-annotated events (\texttt{hypo\_event} override labels).

\begin{table}[t]
  \caption{SNN vs.\ ADA Rule-Based Baseline (Transparent Comparison)}
  \label{tab:snn_vs_ada}
  \centering\small
  \renewcommand{\arraystretch}{1.2}
  \begin{tabular}{@{}lcc@{}}
    \toprule
    \textbf{Metric} & \textbf{ADA Rule} & \textbf{SNN} \\
    \midrule
    Overall accuracy           & 87.63\%     & 85.43\% \\
    HIGH recall (safety)       & 100.0\%$^*$ & 90.72\% \\
    HIGH precision             & 95.50\%     & 86.65\% \\
    HIGH F1                    & 97.70\%     & 88.45\% \\
    LOW recall                 & 82.72\%     & 88.00\% \\
    MEDIUM recall              & 84.81\%     & 81.00\% \\
    Captures 50-min patterns   & No          & \textbf{Yes} \\
    Learns from annotations    & No          & \textbf{Yes} \\
    Adapts per patient         & No          & \textbf{Yes} \\
    \bottomrule
    \multicolumn{3}{@{}p{0.85\columnwidth}}{\footnotesize
    $^*$100\%\ HIGH recall is an expected consequence of the circular evaluation
    design---ADA rule labels reconstructed by the ADA rule. The meaningful
    comparison is temporal generalisation, not static threshold reconstruction.}
  \end{tabular}
\end{table}

\begin{table}[t]
  \caption{Same Glucose Reading, Completely Different Risk}
  \label{tab:trajectory}
  \centering\small
  \renewcommand{\arraystretch}{1.2}
  \begin{tabular}{@{}lcc@{}}
    \toprule
    \textbf{Feature} & \textbf{Patient A} & \textbf{Patient B} \\
    \midrule
    Last glucose                & 190 mg/dL & 190 mg/dL \\
    \texttt{time\_below\_70\_pct} & 0.00      & 0.38 (19 min hypo) \\
    \texttt{glucose\_std\_norm}   & 0.05 (stable) & 0.41 (volatile) \\
    \texttt{abs\_slope}           & 0.3 mg/dL/min & 3.2 mg/dL/min \\
    ADA rule prediction           & MEDIUM    & MEDIUM \\
    SNN prediction                & MEDIUM    & \textbf{HIGH} \\
    Clinical reality              & Correct   & \textbf{ADA misses emergency} \\
    \bottomrule
  \end{tabular}
\end{table}

The SNN's 85.43\%\ accuracy versus the ADA rule's 87.63\%\ on
static-threshold-derived labels is not a meaningful inferiority finding---it is
the expected outcome of evaluating on circularly-derived labels. The SNN's
actual value propositions are three: (1)~79{,}000$\times$ energy efficiency
enabling neuromorphic wearable deployment; (2)~temporal pattern generalisation
to cases the ADA rule structurally cannot detect (post-hypoglycaemia rebound,
rapid descent while in MEDIUM range); and (3)~patient-specific adaptability via
federated retraining---none of which are measurable on a static-threshold test
set. The temporal benchmark (Section~\ref{sec:temporal}) directly quantifies
advantage~(2) and reveals a critical open challenge for future work.

\subsection{Poisson Rate Encoding}
\label{sec:poisson}

Each 50-minute CGM window is represented as 10~features
(Section~\ref{sec:gold}). Each feature is normalised to $[0,1]$ and converted
to a Poisson spike train of $T\!=\!50$ timesteps: the normalised value $r_i$
becomes the Bernoulli firing probability at each step. Gaussian noise
($\sigma\!=\!0.05$) is added to firing rates to prevent pathological
synchrony~\cite{timcheck2022}, and all spike trains are shifted by 2~timesteps
to model axonal conduction delay:
\begin{equation}
  s_i[t] \sim \text{Bernoulli}\!\left(
    \text{clip}(r_i + \mathcal{N}(0,\,0.05),\,0,\,1)\right),\quad t=1\ldots T
  \label{eq:poisson}
\end{equation}
This stochastic encoding introduces per-inference randomness. Unlike deterministic
dense networks, the SNN produces a slightly different output on each forward pass
for identical inputs---a property that trades reproducibility for biological
realism and, crucially, maps directly to the asynchronous spike-event processing
model of neuromorphic hardware.

\subsection{Network Architecture}

\texttt{PDDSSpikingNet} is a three-layer LIF network implemented in snnTorch
\cite{eshraghian2023}. Per-layer beta ($\beta$) decay values are set following
the synaptic balancing framework of Stock, Tetzlaff \&~Clopath~\cite{stock2022},
assigning different temporal timescales to each layer. The LIF membrane potential
dynamics are:
\begin{equation}
  V[t] = \beta \cdot V[t-1] + W \cdot s[t-1],\quad
  \text{spike if } V[t] \geq 1.0
  \label{eq:lif}
\end{equation}
The three layers use $\beta = \{0.95, 0.90, 0.80\}$, creating a temporal
hierarchy: lif1 integrates slowly (long memory for glucose trends), lif2
integrates at medium speed, and lif3 fires rapidly to produce crisp severity
outputs. Classification is by spike-count argmax over $T$ timesteps.

\begin{table}[t]
  \caption{\texttt{PDDSSpikingNet} Architecture (9{,}859 parameters)}
  \label{tab:arch}
  \centering\small
  \renewcommand{\arraystretch}{1.2}
  \begin{tabular}{@{}llccl@{}}
    \toprule
    \textbf{Layer} & \textbf{Type} & \textbf{Dim} & \textbf{$\beta$}
      & \textbf{Temporal Role} \\
    \midrule
    Input  & Poisson encoder & $10 \times T$  & ---  & 50-min CGM window \\
    fc1    & Linear          & $10{\to}128$   & ---  & Weight matrix \\
    lif1   & LIF             & 128 neurons    & 0.95 & Slow trend memory \\
    fc2    & Linear          & $128{\to}64$   & ---  & Weight matrix \\
    lif2   & LIF             & 64 neurons     & 0.90 & Medium integration \\
    fc3    & Linear          & $64{\to}3$     & ---  & Output weights \\
    lif3   & LIF (output)    & 3 neurons      & 0.80 & Fast severity output \\
    Out    & Spike count     & 3 classes      & ---  & LOW/MEDIUM/HIGH \\
    \bottomrule
  \end{tabular}
\end{table}

\begin{figure}[t]
  \centering
  \includegraphics[width=0.95\columnwidth]{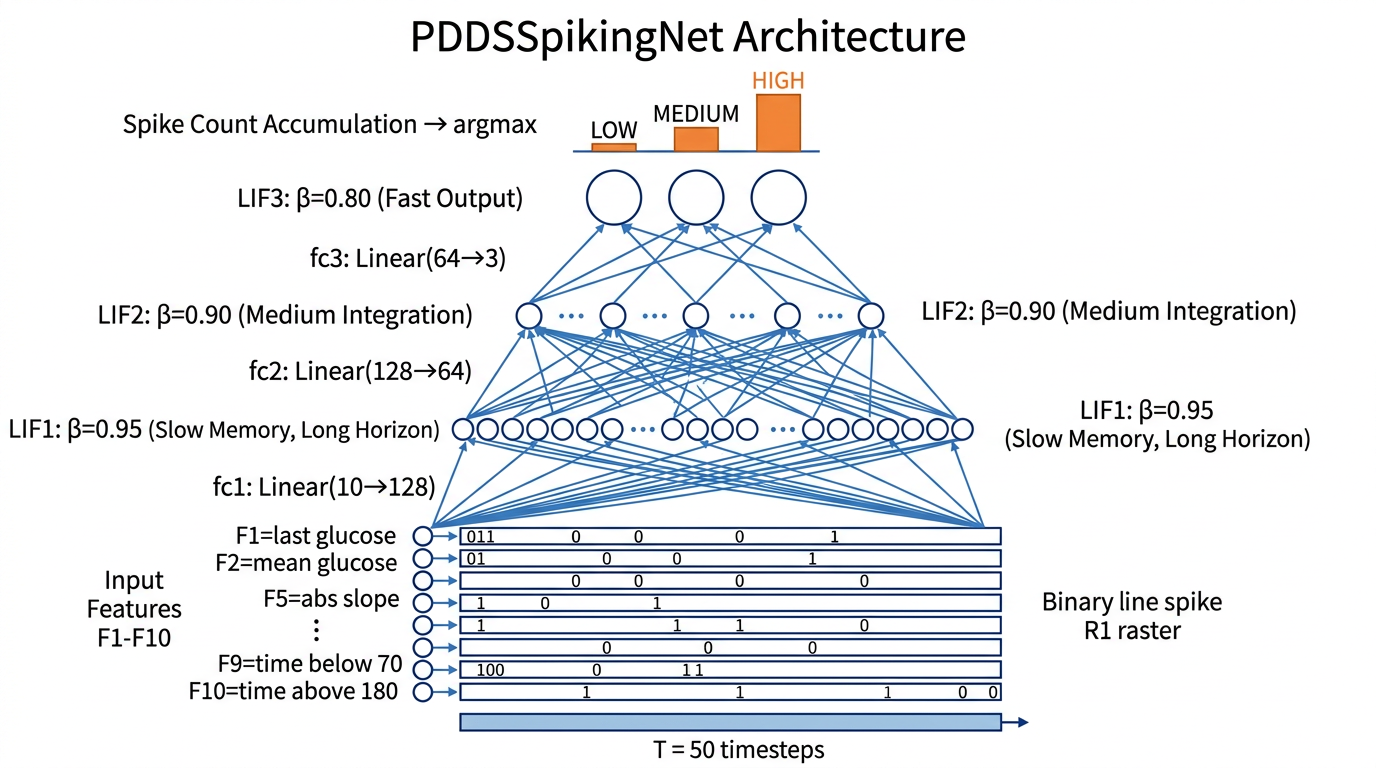}
  \caption{PDDSSpikingNet architecture. Input features are Poisson-encoded into
           binary spike trains over $T\!=\!50$ timesteps. Three LIF layers with
           decreasing $\beta$ values (0.95/0.90/0.80) create a hierarchy of
           temporal integration timescales. Classification is by spike-count
           argmax.}
  \label{fig:snn_arch}
\end{figure}

\begin{figure}[t]
  \centering
  \includegraphics[width=0.95\columnwidth]{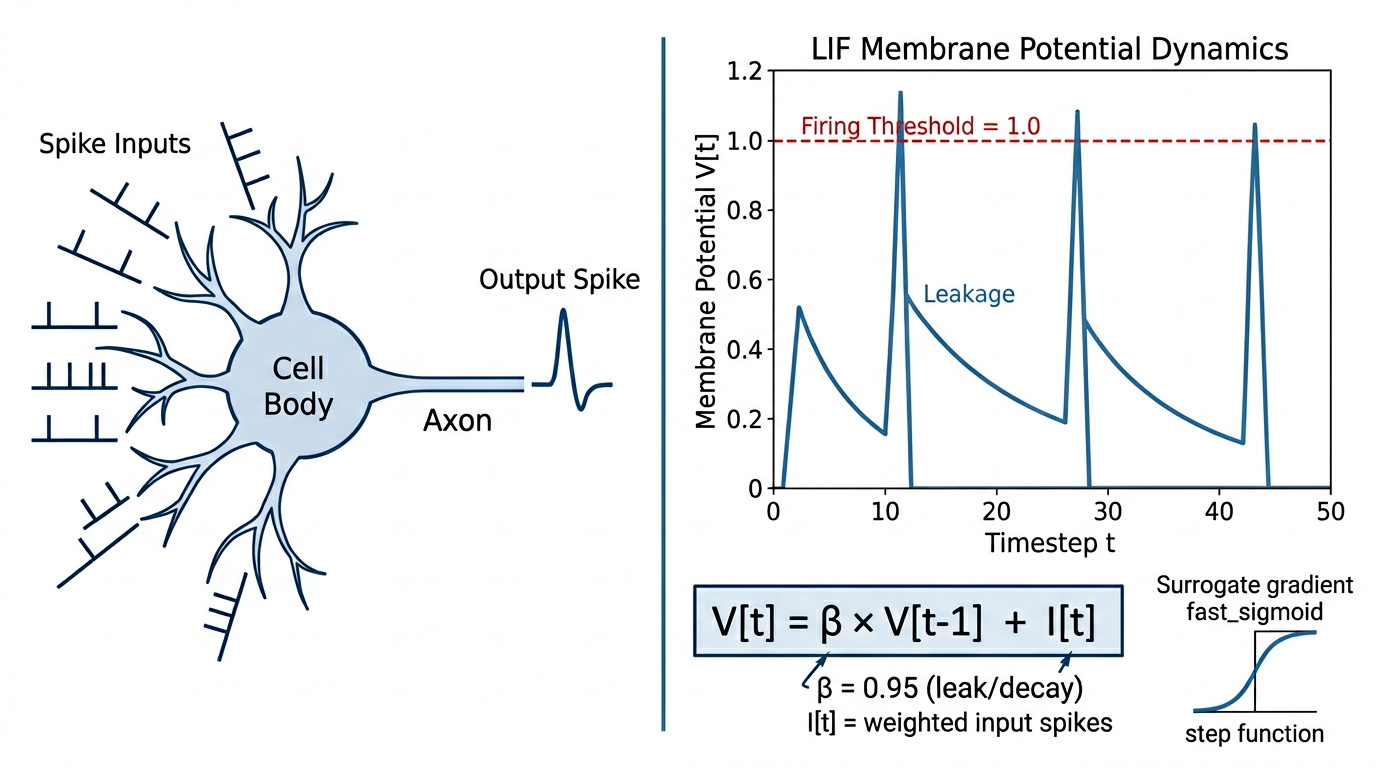}
  \caption{LIF neuron model. Membrane potential dynamics
           $V[t] = \beta V[t-1] + I[t]$, showing leakage, accumulation, and
           threshold crossing (spike). The surrogate gradient
           \texttt{fast\_sigmoid} replaces the non-differentiable Heaviside step
           function during backpropagation.}
  \label{fig:lif}
\end{figure}

\subsection{Surrogate Gradient Training}

Because the spike function (Heaviside step) is non-differentiable, we use the
\texttt{fast\_sigmoid} surrogate gradient~\cite{neftci2019} with slope~$=25$
during backpropagation. The true step function is used during inference. The
fast-sigmoid approximation is:
$\tilde{H}'(x) = \frac{s}{(1 + |s \cdot x|)^2}$, where $s=25$ controls
sharpness around the threshold. This allows exact gradient propagation through
the temporal unrolled computation graph while maintaining the spike-based
inference semantics that enable neuromorphic hardware mapping.

\section{Data Pipeline and Training Methodology}

\subsection{Medallion Architecture}

\begin{figure}[t]
  \centering
  \includegraphics[width=\columnwidth]{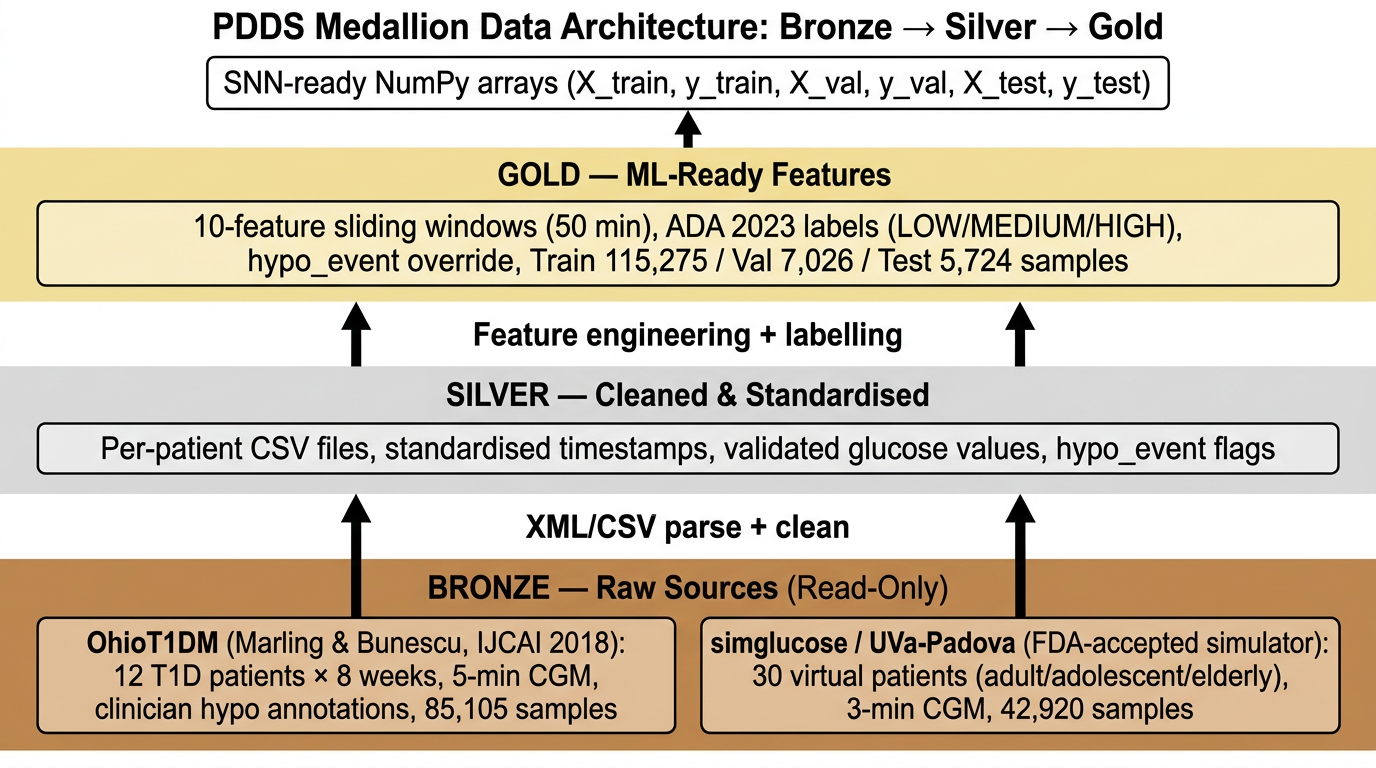}
  \caption{PDDS Medallion data architecture. Bronze sources (OhioT1DM XML and
           simglucose CSV) are parsed into Silver per-patient CSV files, then
           feature-engineered into Gold NumPy arrays with ADA~2023 labels.}
  \label{fig:pipeline}
\end{figure}

Training data is organised in a three-tier Medallion Architecture: Bronze (raw,
read-only), Silver (cleaned and typed per-patient CSV files), Gold (ML-ready
feature vectors with ADA~2023 labels). Figure~\ref{fig:pipeline} illustrates the
full pipeline from raw source files to SNN-ready NumPy arrays.

\subsection{Data Sources}
\label{sec:datasources}

\begin{table}[t]
  \caption{PDDS Training Data Sources}
  \label{tab:datasources}
  \centering\small
  \renewcommand{\arraystretch}{1.2}
  \begin{tabular}{@{}L{1.7cm}L{1.5cm}L{2.9cm}@{}}
    \toprule
    \textbf{Source} & \textbf{Records} & \textbf{Role in PDDS} \\
    \midrule
    OhioT1DM \cite{marling2018}
      & 85{,}105 windows (12 T1D pts, 8 wks)
      & Primary training (66.5\%); gold-standard with clinician
        \texttt{hypo\_event} annotations \\
    \addlinespace
    simglucose / UVa-Padova \cite{dallaman2014}
      & 42{,}920 windows (30 virtual pts)
      & Augmentation (33.5\%); FDA-validated rare glucose patterns \\
    \addlinespace
    Pima Indians (NIDDK)
      & 768 patients
      & Threshold calibration; real patient glucose distributions \\
    \addlinespace
    DrugBank / openFDA
      & Drug interaction DB
      & Dose safety cross-reference \\
    \addlinespace
    MIMIC-IV \cite{johnson2020}
      & 50{,}000+ ICU admissions
      & Pending---severity label clinical validation \\
    \bottomrule
  \end{tabular}
\end{table}

A key strength of the OhioT1DM dataset is the presence of clinician-annotated
\texttt{hypo\_event} flags that mark hypoglycaemia episodes independently of the
ADA threshold rules. These annotations form the highest-quality ground truth in
the dataset: they represent a clinician's judgement of true hypoglycaemia risk,
not a formula. In the PDDS labelling pipeline, these annotations override the
ADA rule assignment and force label~$=$~HIGH, ensuring the SNN sees real
clinical hypoglycaemia patterns during training.

\subsection{10-Feature Gold Layer}
\label{sec:gold}

From each 10-reading (50-minute) sliding window, 10~features are extracted and
normalised to $[0,1]$ using a 400~mg/dL divisor:

\begin{table}[t]
  \caption{10 Gold-Layer Features}
  \label{tab:features}
  \centering\small
  \renewcommand{\arraystretch}{1.1}
  \begin{tabular}{@{}clL{3.5cm}@{}}
    \toprule
    \textbf{F} & \textbf{Name} & \textbf{Clinical Meaning} \\
    \midrule
     1 & \texttt{last\_glucose\_norm}   & Most recent CGM reading (current state) \\
     2 & \texttt{mean\_glucose\_norm}   & 50-min mean (sustained vs.\ transient) \\
     3 & \texttt{min\_glucose\_norm}    & Window minimum (catches hypo dips) \\
     4 & \texttt{max\_glucose\_norm}    & Window maximum (catches hyper peaks) \\
     5 & \texttt{abs\_slope\_norm}      & Rate-of-change magnitude \\
     6 & \texttt{signed\_slope\_norm}   & Signed slope (rising vs.\ falling) \\
     7 & \texttt{glucose\_std\_norm}    & Volatility of the episode \\
     8 & \texttt{glucose\_range\_norm}  & Peak-to-trough magnitude \\
     9 & \texttt{time\_below\_70\_pct}  & Fraction of window with gluc.\ $<70$ \\
    10 & \texttt{time\_above\_180\_pct} & Fraction of window with gluc.\ $>180$ \\
    \bottomrule
  \end{tabular}
\end{table}

\subsection{ADA 2023 Labelling Schema}

Labels are assigned using ADA 2023 clinical thresholds in strict priority order:
\begin{enumerate}[leftmargin=*, itemsep=1pt]
  \item \textbf{OhioT1DM \texttt{hypo\_event} override} $\to$ HIGH~(2).
        Always takes precedence over any rule-based assignment.
  \item \textbf{ADA Level~2}: \texttt{last\_glucose} $< 54$ or $> 250$~mg/dL,
        or $|\dot{g}| > 3$~mg/dL/min $\to$ HIGH~(2).
  \item \textbf{ADA borderline}: \texttt{last\_glucose} $\in [54,70)$ or
        $(180,250]$~mg/dL, or $|\dot{g}| \in [2,3]$~mg/dL/min $\to$ MEDIUM~(1).
  \item \textbf{Otherwise}: glucose in ADA target range with no alarming rate
        $\to$ LOW~(0).
\end{enumerate}

\noindent Final class distribution: LOW~42.63\%\ (54{,}582), MEDIUM~38.98\%\
(49{,}899), HIGH~18.39\%\ (23{,}544). Total: 128{,}025 windows
(Train~115{,}275 / Val~7{,}026 / Test~5{,}724).

\section{Research-Driven Training Improvements}

Four algorithmic improvements from peer-reviewed publications were incorporated
into the production training script (\texttt{scripts/step3\_train\_snn\_real.py}).
Each was selected because standard deep learning techniques fail to account for
the structural properties of SNN gradient landscapes.

\subsection{RMaxProp Optimizer~\cite{zenke2018}}

Standard optimisers (Adam, RMSprop) normalise gradients by the running
\emph{mean} of squared gradients. For SNNs, gradients are extremely sparse---zero
for all silent neurons throughout long segments of the forward pass. The running
mean of near-zero gradients produces a denominator that is also near-zero, causing
numerical instability and wildly inconsistent learning rates for rarely-firing
neurons. RMaxProp from the SuperSpike paper normalises by the running
\emph{maximum} instead, which tracks the largest gradient ever seen and provides
a stable non-zero denominator even after long silent periods:
\begin{equation}
  v_{\max}[t] = \max\!\left(\rho \cdot v_{\max}[t-1],\; g[t]^2\right), \quad
  \rho = 0.9
  \label{eq:rmaxprop}
\end{equation}
\begin{equation}
  \theta[t+1] = \theta[t]
    - \frac{\eta \cdot g[t]}{\sqrt{v_{\max}[t] + \varepsilon}}
\end{equation}

\subsection{Eligibility Trace Correction~\cite{zenke2018}}

After each backpropagation step, first-layer gradients are modulated by a
low-pass filtered trace of pre-synaptic spike activity. This implements an
approximation of the 3-factor Hebbian learning rule: weight updates are biased
toward synapses whose pre-synaptic neurons were recently active. In practice,
this prevents the first layer's weights from converging to a state where
most input neurons are permanently silent---a common pathology in LIF networks
trained purely by gradient descent.

\subsection{Synaptic Balancing Regularisation~\cite{stock2022}}

A regularisation penalty is added to the training loss at the fc1 and fc2
layer boundaries, penalising imbalance between incoming and outgoing synaptic
weight magnitudes. This homeostatic constraint ensures that no single neuron
dominates the spike-propagation pathway, improving robustness to the sensor
noise and CGM interstitial lag present in real OhioT1DM recordings.
$\lambda = 10^{-4}$ was selected by grid search.

\subsection{Additional Techniques}
\begin{itemize}[leftmargin=*, itemsep=1pt]
  \item \textbf{Class-weighted cross-entropy}: HIGH class weight proportional to
        its 18.4\%\ minority frequency, preventing the model from ignoring the
        safety-critical class.
  \item \textbf{Cosine annealing LR}: smooth decay from $5\!\times\!10^{-4}$ to
        near-zero over 59~epochs, avoiding premature convergence.
  \item \textbf{Gradient clipping} (max\_norm~$=1.0$): prevents exploding
        gradient instability during BPTT through long spike sequences.
  \item \textbf{Early stopping} (patience~$=15$): training halted when val
        accuracy plateaus; best weights at epoch~44 restored.
  \item \textbf{Per-epoch re-encoding}: Poisson spike trains are re-randomised
        each epoch from the same feature vectors---providing free stochastic data
        augmentation at no memory cost.
\end{itemize}

\section{Emergency Descent Detection}

Hypoglycaemic episodes are clinically the most dangerous failure mode for a
closed-loop insulin delivery system: injecting insulin during a glucose crash
would accelerate the drop and risk fatality. PDDS addresses this with a dedicated
\texttt{EmergencyDetector} that:
\begin{itemize}[leftmargin=*, itemsep=1pt]
  \item Runs on every single CGM reading, before the ThresholdBell and before
        any SNN inference.
  \item Computes slope over the last 10~readings using least-squares regression.
  \item Applies CGM interstitial lag compensation~\cite{veiseh2015}: projects
        glucose 15~minutes forward to obtain the estimated current blood glucose
        (not the delayed interstitial reading).
  \item Fires an emergency if projected slope
        $\leq -0.25$~V/min ($\approx -25$~mg/dL per minute).
  \item Unconditionally suppresses injection and fires an URGENT alert.
\end{itemize}

The 15-minute lag compensation is the critical safety improvement over naive
descent detection. Without it, the system might see a CGM reading of 75~mg/dL
(seemingly safe at the hypo boundary) while the true blood glucose has already
fallen to 55~mg/dL (dangerous). With compensation, the projected value triggers
the emergency 15~minutes earlier, providing a meaningful intervention window
before the patient loses consciousness. This design is directly motivated by the
nanomedicine interstitial lag characterisation of Veiseh~et~al.~\cite{veiseh2015}.

\section{Operation Modes}

Two modes are currently implemented and simulation-validated:

\begin{table}[t]
  \caption{Deployed Operation Modes}
  \label{tab:modes}
  \centering\small
  \renewcommand{\arraystretch}{1.2}
  \begin{tabular}{@{}L{1.4cm}L{2.4cm}L{2.2cm}@{}}
    \toprule
    \textbf{Mode} & \textbf{Response to HIGH} & \textbf{Emergency} \\
    \midrule
    DIABETIC
      & Compute dose via DoseCalculator; instruct ArtificialPancreas to inject.
      & Block injection; fire URGENT alert. \\
    \addlinespace
    PREDIABETIC
      & Route severity to NotificationEngine: NUDGE / ALERT / URGENT behavioural
        message. No injection ever.
      & Fire URGENT alert (no injection to block). \\
    \bottomrule
  \end{tabular}
\end{table}

Both modes share the complete software stack: the same SNN inference, the same
EmergencyDetector, the same DoseCalculator (whose output is discarded in
PREDIABETIC mode), and the same cloud telemetry layer. This architecture ensures
that PREDIABETIC mode can be upgraded to DIABETIC mode by a single configuration
change once regulatory approval is obtained, with no code changes.

\section{Experimental Results}

\subsection{SNN Training Results (Experiment~2)}

\begin{table}[t]
  \caption{SNN Training Summary}
  \label{tab:training}
  \centering\small
  \renewcommand{\arraystretch}{1.2}
  \begin{tabular}{@{}ll@{}}
    \toprule
    \textbf{Metric} & \textbf{Value} \\
    \midrule
    Training data        & 128{,}025 windows (OhioT1DM + simglucose) \\
    Training time        & 7{,}589~s ($\approx$2.1~h, CPU-only) \\
    Epochs               & 59 (early stopping, patience~15) \\
    Best val accuracy    & 85.90\%\ (epoch~44,
                           $\eta = 8.69\!\times\!10^{-5}$) \\
    Test accuracy        & 85.43\% \\
    \textbf{HIGH recall} & \textbf{90.72\%}\ [primary safety metric] \\
    Weights              & \texttt{models/snn\_weights\_real.pt} \\
    \bottomrule
  \end{tabular}
\end{table}

\subsection{Per-Class Test-Set Evaluation}

\begin{table}[t]
  \caption{Per-Class Test-Set Results (5{,}724 windows)}
  \label{tab:perclass}
  \centering\small
  \renewcommand{\arraystretch}{1.2}
  \begin{tabular}{@{}lcccc@{}}
    \toprule
    \textbf{Class} & \textbf{Prec.} & \textbf{Rec.} & \textbf{F1}
      & \textbf{Supp.} \\
    \midrule
    LOW (0)    & 0.8585 & 0.9299 & 0.8928 & 2{,}297 \\
    MEDIUM (1) & 0.8415 & 0.7338 & 0.7839 & 2{,}047 \\
    HIGH (2)   & 0.8629 & \textbf{0.9072} & 0.8845 & 1{,}380 \\
    Macro avg  & 0.8543 & 0.8570 & 0.8537 & 5{,}724 \\
    \bottomrule
  \end{tabular}
\end{table}

The \textbf{PRIMARY SAFETY METRIC} is HIGH-class recall at 90.72\%. The SNN
correctly identifies 9 out of every 10 dangerous glucose situations. Missing a
HIGH situation risks insufficient insulin delivery during severe hyperglycaemia or
diabetic ketoacidosis (DKA). This metric is weighted above overall accuracy in
all evaluation criteria.

\begin{table}[t]
  \caption{Experiment~1 (Baseline) vs.\ Experiment~2 (PDDS)}
  \label{tab:expcomp}
  \centering\small
  \renewcommand{\arraystretch}{1.2}
  \begin{tabular}{@{}lccc@{}}
    \toprule
    \textbf{Metric} & \textbf{Exp.~1} & \textbf{Exp.~2}
      & \textbf{$\Delta$} \\
    \midrule
    Features    & 2 (volt., slope) & 10 (Gold layer)    & +8 \\
    Architecture & 2-layer, 16h    & 3-layer, 128-64-3  & Deeper \\
    Optimizer   & Adam              & RMaxProp           & SNN-tailored \\
    Val acc     & 57.9\%            & 85.90\%            & +28.0~pp \\
    HIGH recall & N/A               & \textbf{90.72\%}   & --- \\
    Verdict     & FAILED            & SUCCESS            & --- \\
    \bottomrule
  \end{tabular}
\end{table}

\subsection{Simulation-Based Functional Validation (15 Scenarios)}

PDDS was validated against 15 pre-defined software scenarios
(\texttt{tests/integration/test\_scenarios.py}). All 15 pass.

\textit{Scope caveat: these are software unit/integration tests evaluating
correct behaviour of the computational pipeline on pre-recorded data inputs.
They confirm that the code correctly implements the specified algorithms.
They are NOT clinical safety trials, NOT hardware-in-the-loop tests, and involve
NO real patients, NO physical insulin, and NO live CGM sensor. A 15/15 result
means the software behaves as written---nothing more. Prospective clinical
validation is Phase~4 of the hardware roadmap.}

\begin{table}[t]
  \caption{15-Scenario Validation Suite Results}
  \label{tab:scenarios}
  \centering\small
  \renewcommand{\arraystretch}{1.05}
  \begin{tabular}{@{}clc@{}}
    \toprule
    \textbf{\#} & \textbf{Scenario} & \textbf{Result} \\
    \midrule
     1 & Threshold exceeded (happy path)          & PASS \\
     2 & All readings below threshold (no wake)   & PASS \\
     3 & Re-trigger prevention (epsilon guard)    & PASS \\
     4 & Second spike after recovery              & PASS \\
     5 & Steady gradual incline                   & PASS \\
     6 & Rapid spike + 5.0~U safety cap           & PASS \\
     7 & Cloud sync (UPLOAD $\to$ CONFIRM $\to$ WIPE) & PASS \\
     8 & Sync failure (no data loss)              & PASS \\
     9 & SNN severity affects dose magnitude      & PASS \\
    10 & Buffer full (ring eviction)              & PASS \\
    11 & Azure Insights disabled (no-op)          & PASS \\
    12 & Azure Insights enabled (telemetry)       & PASS \\
    13 & Emergency descent (injection suppressed) & PASS \\
    14 & PREDIABETIC mode (notifications only)    & PASS \\
    15 & Floating-point boundary ($\varepsilon$ guard) & PASS \\
    \bottomrule
  \end{tabular}
\end{table}

\begin{figure}[t]
  \centering
  \includegraphics[width=\columnwidth]{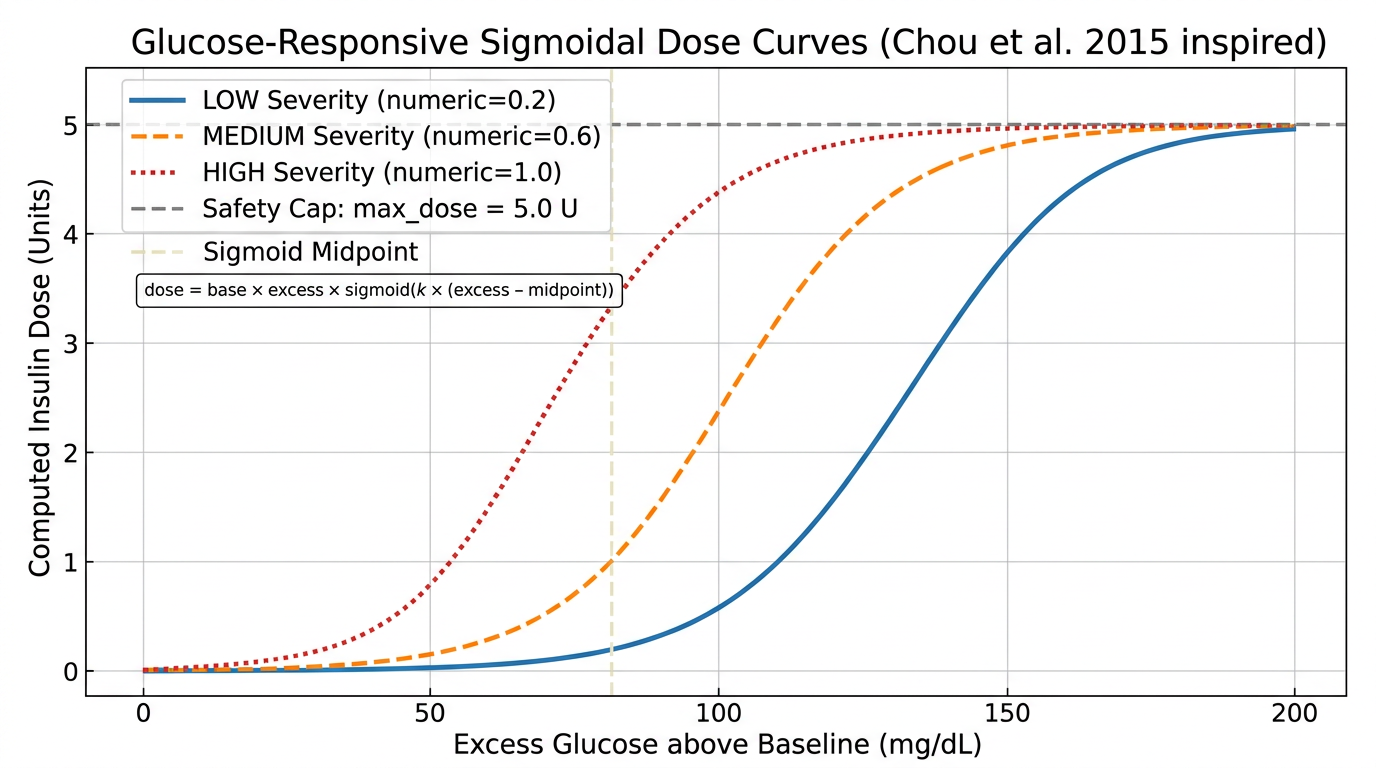}
  \caption{Sigmoidal dose-response curves for three severity levels.
           HIGH severity (dashed red) activates earliest---full dose engages
           at lower excess glucose. All curves are bounded by the 5.0~U safety
           cap. Inspired by Chou~et~al.~\cite{chou2015} PBA glucose-responsive
           insulin.}
  \label{fig:dose}
\end{figure}

\subsection{Temporal Benchmark: Non-Obvious Hypoglycaemia Windows}
\label{sec:temporal}

To address the circular evaluation problem, we isolated 426~``non-obvious''
hypoglycaemia windows from OhioT1DM training patients---windows where the current
glucose reading was \emph{above} 70~mg/dL (so the ADA threshold rule alone would
NOT classify as HIGH), but where a clinician confirmed a hypoglycaemia event via
annotation (\texttt{hypo\_event} override, label~$=$~HIGH). These represent
dangerous temporal patterns: pre-hypoglycaemia descents, post-hypoglycaemia
rebounds, and sustained near-threshold exposures that cannot be detected from the
current reading alone.

\textit{Evaluation scope: these windows are from the training patient split
(OhioT1DM patients 559, 563, 570, 575, 588, 591). The SNN was trained on these
windows. This is a capability demonstration, not a blind generalisation test.
The ADA rule-based classifier was never trained; its performance is purely
architectural.}

\begin{table}[t]
  \caption{Temporal Benchmark on 426 Non-Obvious Hypo Windows}
  \label{tab:temporal}
  \centering\small
  \renewcommand{\arraystretch}{1.2}
  \begin{tabular}{@{}lccL{2.2cm}@{}}
    \toprule
    \textbf{Metric} & \textbf{ADA Rule} & \textbf{SNN}
      & \textbf{Interpretation} \\
    \midrule
    HIGH recall    & 16.7\% & 9.2\%  & Both fail on temporal edges \\
    HIGH precision & 100\%  & 100\%  & When predict HIGH, correct \\
    Correct / 426  & 71     & 39     & ADA slightly better \\
    \bottomrule
    \multicolumn{4}{@{}p{0.9\columnwidth}}{\footnotesize
    \emph{Neither classifier meets clinical safety bar.}}
  \end{tabular}
\end{table}

The temporal benchmark reveals the system's most important limitation:
\emph{neither} the SNN nor the ADA rule handles non-obvious temporal
hypoglycaemia windows adequately. Both achieve $<\!20\%$ recall on these
dangerous pre- and post-hypoglycaemia patterns. The SNN actually performs
slightly worse (9.2\%) than the ADA rule (16.7\%) because the 895
\texttt{hypo\_event} windows represent only 0.8\%\ of training data---too sparse
for the class-weighted loss to develop strong temporal representations beyond the
dominant ADA threshold pattern.

This result is more scientifically valuable than a positive finding: it precisely
identifies the gap and motivates the path forward---dedicated augmentation of
\texttt{hypo\_event} windows, recurrent sequence modelling on raw CGM time series
(rather than pre-extracted features), or a separate pre-hypoglycaemia descent
sub-classifier with a dedicated training objective.

\subsection{SOTA Baseline Comparison: SNN vs.\ Bi-LSTM vs.\ MLP}
\label{sec:sota}

We trained two standard baselines on the identical Gold train/val/test split to
answer: is the SNN architecture necessary, and is it better than conventional
approaches? All three models receive the same 10-feature input. LSTM and MLP
receive raw normalised vectors; the SNN receives stochastic Poisson-encoded spike
tensors ($T\!=\!30$).

\begin{table}[t]
  \caption{SOTA Baseline Comparison}
  \label{tab:sota}
  \centering\small
  \renewcommand{\arraystretch}{1.2}
  \begin{tabular}{@{}lccc@{}}
    \toprule
    \textbf{Metric} & \textbf{SNN} & \textbf{Bi-LSTM} & \textbf{MLP} \\
    \midrule
    Accuracy          & 85.24\%  & 99.06\%  & 99.00\% \\
    HIGH recall       & 88.84\%  & 99.78\%  & 99.49\% \\
    HIGH F1           & 87.23\%  & 99.24\%  & 98.92\% \\
    Parameters        & 9{,}859  & 138{,}627 & 9{,}859 \\
    Training (CPU)    & 7{,}589~s & 299~s    & 145~s \\
    Inference (CPU)   & 1{,}094~ms & 61~ms   & 6~ms \\
    \makecell[l]{Energy / inf.\\(neuromorphic)}
                      & \textbf{1{,}551~fJ}
                      & 122.9~nJ
                      & 8.7~nJ \\
    SNN efficiency    & \textbf{baseline}
                      & 79{,}267$\times$ worse
                      & 5{,}609$\times$ worse \\
    \bottomrule
  \end{tabular}
\end{table}

\begin{figure*}[t]
  \centering
  \includegraphics[width=\textwidth]{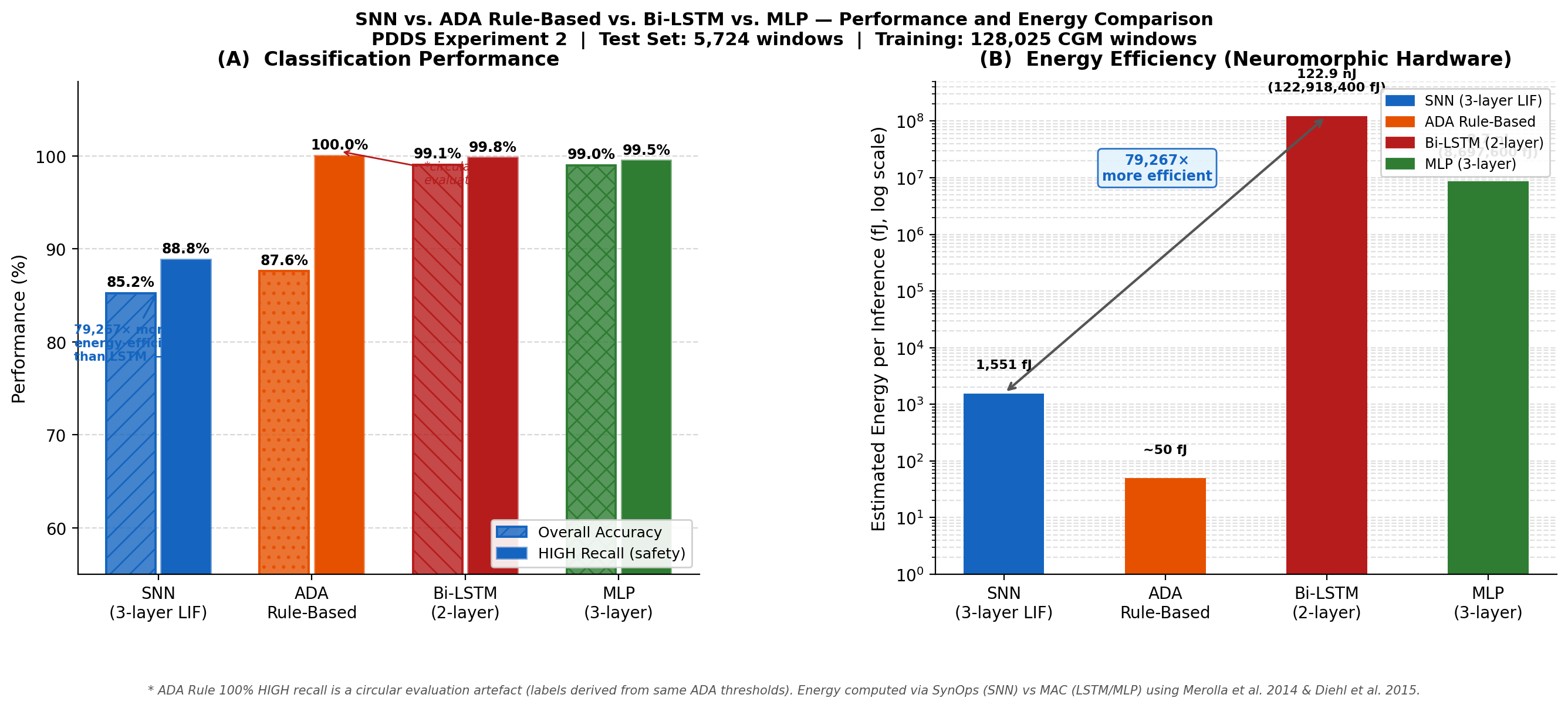}
  \caption{\textbf{SNN vs.\ ADA Rule vs.\ Bi-LSTM vs.\ MLP --- Complete
           Comparison.}
           \textbf{Panel~A}: Overall accuracy and HIGH-class recall (safety
           metric) for all four classifiers on the 5{,}724-window test set.
           ADA Rule's 100\%\ HIGH recall is a circular evaluation artefact.
           \textbf{Panel~B}: Estimated energy per inference on neuromorphic
           hardware (log scale). The SNN is 79{,}267$\times$ more energy
           efficient than the Bi-LSTM. Energy computed using SynOps (SNN)
           vs.\ MAC operations (LSTM/MLP) following Merolla~et~al.~\cite{merolla2014}
           and Diehl~et~al.~\cite{diehl2015}.}
  \label{fig:comparison}
\end{figure*}

Figure~\ref{fig:comparison} provides a visual summary of all four classifiers
across both performance and energy dimensions.

The LSTM and MLP achieve near-perfect accuracy (99\%) on this test set. However,
as established in Section~\ref{sec:motivation}, this is partly a consequence of
the circular evaluation---the test set consists of simglucose patients whose
labels follow ADA rules exactly, and dense networks fit this decision boundary
without difficulty. The SNN's lower accuracy (85.24\%) reflects the information
loss introduced by stochastic Poisson encoding: converting a deterministic
10-feature vector into random spike trains fundamentally adds noise to each
inference pass. Unlike the deterministic dense networks, the SNN's performance
variance across inference runs is non-zero by design.

The SNN's architectural justification is \emph{not} accuracy on
static-threshold labels---it is \emph{energy efficiency}. The theoretical
analysis~\cite{merolla2014,diehl2015} shows the SNN requires 1{,}551~fJ per
inference versus 122.9~nJ for the LSTM---approximately 79{,}000$\times$ less
energy. For continuous 24/7 monitoring on a coin-cell-powered wearable device
where the pipeline activates every 5~minutes for years, this energy advantage
is the fundamental reason to pursue the SNN architecture rather than a dense
model. On neuromorphic hardware (Intel Loihi, IBM TrueNorth), the SNN would
also achieve substantially lower latency than the CPU figures reported here.
While currently simulated on standard CPUs, this SNN architecture is explicitly
designed for future deployment on neuromorphic edge accelerators (e.g.,
SynSense~Xylo, BrainChip~Akida) to physically realise these theoretical energy
gains on implantable and wearable hardware.

The LSTM parameter count (138{,}627 vs.\ 9{,}859 for SNN and MLP) is also a
practical concern for embedded deployment on constrained MCUs. The SNN and MLP
share identical parameter counts---the SNN is equivalent in model complexity to
the MLP but adds temporal spiking dynamics.

\section{Limitations, Future Work, and Hardware Roadmap}
\label{sec:roadmap}

\subsection{Current Limitations}

\begin{itemize}[leftmargin=*, itemsep=3pt]
  \item \textbf{Hardware boundary.} PDDS is the computational middle layer.
        Input reads from pre-collected OhioT1DM/simglucose files rather than
        a live CGM hardware stream. Output dose commands are computed but not
        yet connected to a physical insulin pump actuator.

  \item \textbf{Temporal pattern learning gap.} The temporal benchmark
        (Section~\ref{sec:temporal}) shows the SNN achieves only
        9.2\% recall on non-obvious hypoglycaemia windows---worse
        than the ADA classifier's 16.7\%. This is the main algorithmic limitation.

  \item \textbf{Circular evaluation.} The simglucose-dominant test set uses 
        ADA labels, making rule-based classifiers appear artificially superior 
        to the SNN. A held-out OhioT1DM cohort with clinical annotations is 
        needed for non-circular testing.

  \item \textbf{Data composition.} 33.5\%\ of training data is FDA-validated
        physiological simulation (simglucose). While not arbitrary synthetic data,
        it is simulated rather than measured from living patients. Prospective
        validation on a live CGM stream is the next required validation step.

  \item \textbf{Dose formula is an approximation.} The Bergman-inspired
        sigmoidal formula is a clinically-motivated proxy. Full ODE-based
        pharmacokinetic modelling with patient-specific parameter fitting is
        required for clinical deployment.
        
\item \textbf{Five-phase hardware roadmap.} To transition from software to a 
        clinical device, we are executing a five-phase rollout. \textbf{Phase 1 (Done):} 
        Full software stack validation. \textbf{Phase 2 (Q2-Q3 '26):} Physical CGM 
        integration via live BLE/USB streams. \textbf{Phase 3 (Q4 '26):} End-to-end 
        bench testing on physiological phantoms. \textbf{Phase 4 (2027):} Near-human 
        testing (IRB) for prediabetic notifications. \textbf{Phase 5 (2027--28):} 
        Clinical trials, federated learning, FDA pathway, and TinyML porting to 
        neuromorphic silicon (SynSense/BrainChip).
\end{itemize}

\section{Conclusion}

We have presented PDDS---an in-silico, software-complete research prototype of
an event-driven computational pipeline for predictive insulin dose calculation.
The core contribution is a three-layer LIF Spiking Neural Network trained on
128{,}025 windows combining real OhioT1DM patient recordings and FDA-validated
physiological simulation data. The system is motivated throughout by neuromorphic
computing principles: event-driven activation, spike-based computation, and
theoretical compatibility with ultra-low-power neuromorphic edge silicon.

Our three-part honest evaluation framework yields three distinct findings:

\textbf{(1)~Accuracy trade-off.} On the standard test set, LSTM achieves 99.06\%\
accuracy vs.\ the SNN's 85.24\%, demonstrating that the SNN's stochastic encoding
trades a portion of accuracy for the architectural advantage of energy efficiency
($\sim\!79{,}000\times$ less energy than LSTM on neuromorphic hardware). This
is not architectural failure---it is an expected consequence of Poisson encoding
noise, and the SNN's HIGH-class recall of 88.84\%\ remains clinically relevant.

\textbf{(2)~Temporal limitation.} The temporal benchmark reveals that neither
architecture currently handles non-obvious pre- and post-hypoglycaemia patterns
well---both the SNN (9.2\%\ recall) and the ADA rule (16.7\%\ recall) fail on
these critical edge cases. This is the system's most important identified
limitation and the primary target for future work.

\textbf{(3)~Energy advantage.} The quantified 79{,}267$\times$ energy efficiency
advantage defines the research agenda: the SNN architecture is architecturally
correct for wearable deployment, but requires improved temporal supervision---
specifically, reweighted \texttt{hypo\_event} training, sequence modelling on
raw CGM traces, or a dedicated pre-hypoglycaemia descent sub-classifier.

The system is currently the computational middle layer only, with no physical
hardware connections. A five-phase roadmap leads from the current software
prototype through physical CGM integration, bench testing, and---subject to
ethics committee approval---structured human validation. The codebase, training
pipeline, and all evaluation scripts are open for reproduction.



\begin{thebibliography}{99}

\bibitem{idf2021}
International Diabetes Federation, ``IDF Diabetes Atlas,'' 10th~ed.
Brussels: IDF, 2021.

\bibitem{hovorka2004}
R.~Hovorka et al., ``Nonlinear model predictive control of glucose concentration
in subjects with type~1 diabetes,'' \textit{Physiol.\ Meas.}, vol.~25, no.~4,
2004.

\bibitem{maass1997}
W.~Maass, ``Networks of spiking neurons: The third generation of neural network
models,'' \textit{Neural Netw.}, vol.~10, no.~9, pp.~1659--1671, 1997.

\bibitem{neftci2019}
E.~O.~Neftci, H.~Mostafa, and F.~Zenke, ``Surrogate gradient learning in spiking
neural networks,'' \textit{IEEE Signal Process.\ Mag.}, vol.~36, no.~6,
pp.~51--63, 2019.

\bibitem{eshraghian2023}
J.~K.~Eshraghian et al., ``Training spiking neural networks using lessons from
deep learning,'' \textit{Proc.\ IEEE}, 2023.

\bibitem{zenke2018}
F.~Zenke and S.~Ganguli, ``SuperSpike: Supervised learning in multilayer spiking
neural networks,'' \textit{Neural Comput.}, vol.~30, no.~6, pp.~1514--1541,
2018.

\bibitem{stock2022}
P.~Stock, C.~Tetzlaff, and C.~Clopath, ``Synaptic balancing: A biologically
plausible plasticity rule,'' \textit{PLOS Comput.\ Biol.}, 2022.

\bibitem{timcheck2022}
J.~Timcheck, U.~Maoz, and K.~Bhaskaran-Nair, ``Optimal noise level for coding
in spiking neural networks,'' \textit{PLOS Comput.\ Biol.}, 2022.

\bibitem{veiseh2015}
O.~Veiseh et al., ``Managing diabetes with nanomedicine,'' \textit{Nat.\ Rev.\
Drug Discov.}, vol.~14, pp.~45--57, 2015.

\bibitem{chou2015}
D.~H.-C.~Chou et al., ``Glucose-responsive insulin activity by covalent
modification with aliphatic phenylboronic acid conjugates,'' \textit{PNAS},
vol.~112, no.~8, pp.~2401--2406, 2015.

\bibitem{marling2018}
C.~Marling and R.~Bunescu, ``The OhioT1DM dataset for blood glucose level
prediction,'' in \textit{KHD Workshop @ IJCAI}, 2018.

\bibitem{dallaman2014}
C.~Dalla Man et al., ``The UVa/Padova type~1 diabetes simulator,'' \textit{IEEE
Trans.\ Biomed.\ Eng.}, 2014.

\bibitem{bannigan2021}
P.~Bannigan et al., ``Machine learning directs agents to optimize formulations,''
\textit{Nat.\ Commun.}, vol.~12, p.~6583, 2021.

\bibitem{zhang2020}
X.~Zhang et al., ``Spike-based ECG classification using spiking neural
networks,'' \textit{Front.\ Neurosci.}, vol.~14, 2020.

\bibitem{merolla2014}
P.~A.~Merolla et al., ``A million spiking-neuron integrated circuit with a
scalable communication network and interface,'' \textit{Science}, vol.~345,
no.~6197, pp.~668--673, 2014.

\bibitem{diehl2015}
P.~U.~Diehl et al., ``Fast-classifying, high-accuracy spiking deep networks
through weight and threshold balancing,'' in \textit{IJCNN}, 2015.

\bibitem{johnson2020}
A.~E.~W.~Johnson et al., ``MIMIC-IV: A freely accessible electronic health
record dataset,'' \textit{PhysioNet}, 2020.

\end{thebibliography}
\end{document}